\documentclass[a4paper,twoside]{article}

\usepackage{graphicx}
\graphicspath{{./graphics/}}
\usepackage[utf8]{inputenc}
\usepackage{multirow}

\usepackage{url}
\usepackage{hyperref}
\renewcommand{\figureautorefname}{Fig.}

\usepackage{epsfig}
\usepackage{subcaption}
\usepackage{calc}
\usepackage{amssymb}
\usepackage{amstext}
\usepackage{amsmath}
\usepackage{amsthm}
\usepackage{multicol}
\usepackage{pslatex}

\usepackage[backend=biber, 
style=apa, 
natbib=true, 
hyperref=true, 
]{biblatex}

\usepackage[american]{babel}
\usepackage{csquotes}

\usepackage{SCITEPRESS}     

\newcommand{\Autoref}[1]{%
  \begingroup%
  \renewcommand\figureautorefname{Figure}%
  \autoref{#1}%
  \endgroup%
  }

\begin{document}

\title{Identifying Scenarios in Field Data to Enable Validation of Highly Automated Driving Systems}

\author{\authorname{Christian~Reichenbächer\sup{1; 2}\orcidAuthor{0000-0002-0907-3287}, Maximilian~Rasch\sup{1; 3}\orcidAuthor{0000-0002-7554-2619}, Zafer~Kayatas\sup{1; 3}\orcidAuthor{0000-0003-0880-0304}, Florian~Wirthmüller\sup{1; 4}\orcidAuthor{0000-0002-9732-2561}, Jochen~Hipp\sup{1}\orcidAuthor{0000-0002-9037-9899}, Thao~Dang\sup{5}\orcidAuthor{0000-0001-5505-8953} and Oliver~Bringmann\sup{2}\orcidAuthor{0000-0002-1615-507X}}
\affiliation{\sup{1}Mercedes-Benz AG, Sindelfingen, Germany}
\affiliation{\sup{2}Department of Computer Science, University of Tübingen, Tübingen, Germany}
\affiliation{\sup{3}Institute of Technical Mechanics and Vehicle Dynamics, Brandenburg University of Technology, Cottbus, Germany}
\affiliation{\sup{4}Institute of Databases and Information Systems (DBIS), Ulm University, Ulm, Germany}
\affiliation{\sup{5}Faculty Computer Science and Engineering, Esslingen University of Applied Sciences, Esslingen, Germany}
\email{christian.reichenbaecher@mercedes-benz.com, maximilian.rasch@mercedes-benz.com, zafer.kayatas@mercedes-benz.com, florian.wirthmueller@mercedes-benz.com, jochen.hipp@mercedes-benz.com, thao.dang@hs-esslingen.de, oliver.bringmann@uni-tuebingen.de}
}


\abstract{Scenario-based approaches for the validation of highly automated driving functions are based on the search for safety-critical characteristics of driving scenarios using software-in-the-loop simulations. This search requires information about the shape and probability of scenarios in real-world traffic. The scope of this work is to develop a method that identifies predefined logical driving scenarios in field data, so that this information can be derived subsequently. More precisely, a suitable approach is developed, implemented and validated using a traffic scenario as an example. The presented methodology is based on qualitative modelling of scenarios, which can be detected in abstracted field data. The abstraction is achieved by using universal elements of an ontology represented by a domain model. Already published approaches for such an abstraction are discussed and concretised with regard to the given application. By examining a first set of test data, it is shown that the developed method is a suitable approach for the identification of further driving scenarios.}

\onecolumn \maketitle \normalsize \setcounter{footnote}{0} \vfill

\section{\uppercase{Introduction}}
\label{sec:intro}

\begin{figure}[t!]
\centering\includegraphics[width=1.0\columnwidth]{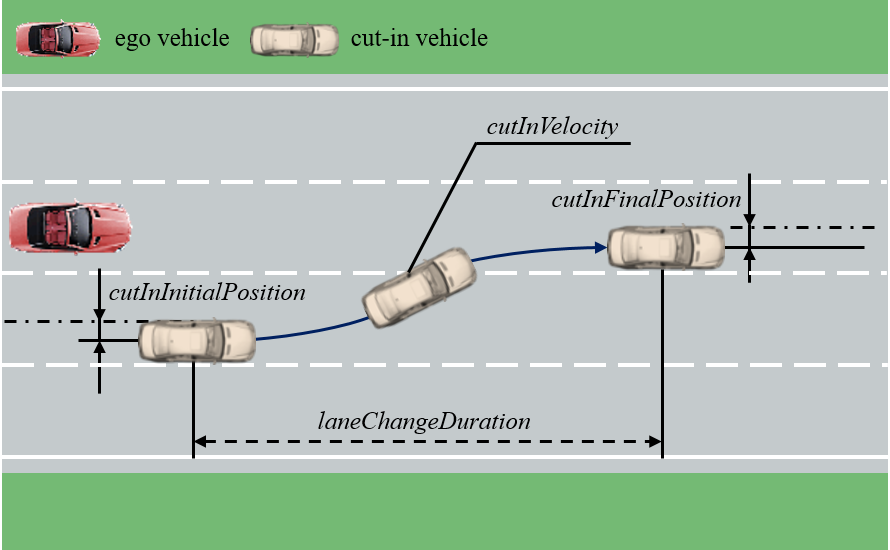}\caption{Trajectory and simulation parameters for the scenario cut-in.}\label{fig:fig1}
\end{figure}

The development of highly automated driving is gaining more and more momentum and is about to find its way into production vehicles soon. However, this technology raises new challenges for the system validation required for series approval. The increased interconnection of driving functions and the replacement of human drivers as a control instance expands the range of requirements and test cases to be validated and verified. Distance-based methods proving adequate system safety that have been used so far, are too costly from an economic point of view. \textcite[p.~458]{Maurer.2015} confirm this fact by determining the test kilometers to be completed for automated driving on the highway in order to prove that the system is at least twice as safe as a human driver, with at least 6.62 billion kilometers.

A new and uniform validation concept was therefore developed as part of the joint project PEGASUS (see \href{https://www.pegasusprojekt.de/en/}{\mbox{pegasusprojekt.de/en}}). The project was conducted in cooperation with automotive manufacturers, suppliers, research institutions and the German Federal Ministry for Economic Affairs and Energy. The developed concept enables to proof the safety of highly automated driving systems with an economically justifiable effort. The methodology derived from this concept for the highway domain in \parencite{rasch} consists of a multi-stage process. In this process, the space of test cases is searched for driving scenarios such as a close cut-in, the end of a traffic jam or other scenarios that are particularly safety-critical. The search for such scenarios is carried out with the help of software-in-the-loop simulations. Initially, common scenarios in the highway domain are logically described on the basis of parameters. These scenarios are among others: following a car in the same lane, a cut-in or a cut-out in front of a vehicle.

Logical scenarios qualitatively describe the basic behaviour of vehicles \parencite{King.2017}. The latter can occur in different varieties in reality, for example at different speeds or distances between the road users involved. Correspondingly quantitative descriptions are called concrete scenarios. A logical scenario thus comprises concrete scenarios with different characteristics of the same basic behaviour of road users.

{\Autoref{fig:fig1}} shows the logical scenario cut-in and visualises some of the simulation parameters. The logical scenario is defined as the description of all scenarios in which a vehicle, here called cut-in vehicle, changes from a lane next to the ego vehicle to the latter's lane. The ego vehicle, shown in red in \autoref{fig:fig1}, is the test object with the highly automated driving functions to be validated. Since its system behaviour is to be examined, its position and state of movement are not specified or parameterised.

By varying the parameters in individual simulation runs, the parameter space of test cases can be searched for critical scenarios in which the system fails. However, even the low-cost software-in-the-loop simulation of all possible variants compared to hardware-in-the-loop simulation, test site or field drives does not make sense from an economic point of view. Thus, those scenario parameterisations should preferably be simulated, which have a sufficiently high probability of occurrence in real-world traffic. In order to determine the probabilities of various scenarios, information about realistic parameter distributions are required.

In this paper a procedure to identify parameterised scenarios in field data is developed. It enables to extract parameter distributions in future. In the course of this, an approach is developed that allows to reduce the field data, describing driving condition and vehicle environment, recorded by the vehicle sensors in an object-oriented manner, to relevant variables. This approach is implemented and validated using the driving scenario Cut-in as an example. In the long term, the concept of this procedure will enable the identification of an entire selection of highway scenarios in field data.

The remainder of this work is structured as follows: In Section \ref{sec:rel_work}, the current state of research on the identification of driving scenarios is presented. In particular, existing approaches for the abstraction of scenarios and measurement data by means of metamodelling are discussed. Section \ref{sec:methods} explains the approach chosen for the identification of scenarios in the context of this work, namely metamodelling with pattern recognition. In detail, definitions are specified and an ontology represented by a domain model is introduced for the metamodelling. Subsequently, the logic of the developed procedure for the abstraction of measurement data and the identification of driving scenarios is described. Section \ref{sec:results} presents the procedure for validating the implemented method and the results obtained. In addition, the results and the methodology of the identification procedure are evaluated. Section \ref{sec:conclusion} summarises the core results of the work, highlights their significance and provides a brief outlook on further research.

\section{\uppercase{Related Work}}
\label{sec:rel_work}
The identification of logical driving scenarios in field data has already been subject to various research. \textcite{King.2017} present an approach for the identification of individual driving maneuvers and logical scenarios in a - in contrast to this work - virtual test drive. According to the authors, driving maneuvers or complex interactions between road users can only be read with difficulty from just the speed and position information of the individual vehicles. For this reason, they consider a reduction and abstraction of the recorded data for interpretation to be mandatory. In this context, they present their approach of using knowledge-based metamodels of scenarios for pattern recognition. For the modelling of the scenarios, they propose to use concept of \textcite{Bach.2016}, which enables an abstraction of real-world driving scenarios down to the logical level. 

The concept is based on a domain model for mapping the relevant classes of a driving scenario. \textcite{Bach.2016} use terminology from the field of theater and film and refer to \textcite{Kienle.2014}. \textcite{Bach.2016} consider a scenario as a specific timespan with a postulated sequence of events. According to the authors, the principle of metamodelling lies in the abstraction of all relevant elements within a logical scenario. The term participant is used to describe all dynamic elements within a scenario that can interact with one another. In order to specify the state and action of the participants on an abstract level, \textcite{Bach.2016} refer to a selection of maneuvers. By assigning a specific sequence of maneuvers, the behaviour of the participants can be modeled within a scenario. A temporal abstraction occurs through so-called acts. Individual acts are distinguished from each other by the maneuvers of the road users. Accordingly, a scenario is defined as a sequence of specific acts that differ precisely in one maneuver by the participants.

For identifying logical scenarios, the StreetWise methodology relies on the use of knowledge-based metamodels for pattern recognition as well \parencite{ElrofaiH.B.H..2018}. With the StreetWise methodology the Dutch Organization for Applied Scientific Research aims to provide a data-based method for the realistic generation of test cases in order to enable the validation and development of automated driving functions. 

For the metamodelling of scenarios, \textcite{ElrofaiH.B.H..2018} revised ontologies from \textcite{Kienle.2014,Ulbrich.2015,elr.2016}, further specified them and created a domain model that is supposed to meet the practical requirements of scenario identification. \textcite{ElrofaiH.B.H..2018} speak of activities instead of manoeuvres and concretises their definition as the temporal change of status variables such as speed and orientation in order to describe, for example, a lane change or braking to a standstill. \textcite{Paardekooper.2019} distinguish between possible longitudinal activities with accelerating, cruising and decelerating as well as lateral activities with lane keeping, lane change to the left and lane change to the right. If at least two of the six activities (longitudinal and lateral) are always assigned to every road user, it should be possible to describe every possible movement trajectory.

\textcite{ElrofaiH.B.H..2018} claim the StreetWise pattern recognition algorithm is supposed to use artificial intelligence methods such as machine learning. \textcite{Paardekooper.2019} adds that a method of "template matching on a graphical network" is used. 

For the description of logical scenarios, \textcite{Gelder.30.01.2020} introduce so-called scenario categories in combination with element categories for activities, static and dynamic environment. To specify a scenario category, the authors suggest to reference the elements that the scenario contains, for example, actors, activities, static environment and events. By doing so, these elements can be used for various scenarios. An activity is described as the temporal change in one or more status variables of an actor between two events. Like \textcite{Paardekooper.2019}, \textcite{Gelder.30.01.2020} suggest a distinction between longitudinal and lateral activities. The latter defines an event as a point in time at which either a defined threshold value is reached or a mode change takes place through which the change in one or more status variables is described by a different behaviour respectively a different activity.

In summary, a method for pattern recognition and its application to real-world data has not yet been described in detail in the research work presented. In this aspect, our work differs from the previous ones.

\section{\uppercase{Methods}}
\label{sec:methods}

The approach chosen in this work for the identification of driving scenarios builds on knowledge-based metamodelling of scenarios in connection with pattern recognition as suggested by \textcite{King.2017}, \textcite{ElrofaiH.B.H..2018} and \textcite{Paardekooper.2019} (see Section \ref{sec:rel_work}). In order to be able to clearly identify and differentiate between logical driving scenarios, an equally clear definition of the latter is necessary. Knowledge-based metamodels are a suitable means for such definitions.

The explicit modelling of the scenarios sought has the advantage that their recognition in field data can be derived directly and is also fully traceable compared to the use of, for example, neural networks. Furthermore, in contrast to machine learning based approaches, large amounts of training data are not necessary for the implementation.

\subsection{Nomenclature and Definitions for Driving Scenarios}\label{sec:definitions}

The term scenario is used in a wide variety of areas, which is why there is no generally valid, but a large number of different definitions for this term. The definition of a logical scenario and related terms in this work largely follow the ones presented in Section \ref{sec:rel_work}, but are further specified or simplified with regard to the present application. This Section summarises the concepts and terms scenario, ego vehicle, actor, act, activity and event used in this work: 

\paragraph{Scenario}

In general, a scenario is defined as a postulated sequence of events with certain activities and properties of the ego vehicle and its static and dynamic environment. 

\paragraph{Ego Vehicle}

The ego vehicle relates to the perspective from which the world is seen. More precisely, the ego vehicle refers to the vehicle that perceives the environment through the built-in sensors. This includes the driver and optional automated driving functions. 

\paragraph{Actor}

Actors are all dynamic elements of a scenario that are able to interact with each other. The ego vehicle is also assigned to the category actor, as it has the same characteristics as other vehicles. Here, the term actor does not presuppose an actual movement or interaction of the elements, but merely requires their ability to do so in principle. 

\paragraph{Act}

The definition of acts enables a temporal abstraction of the driving events. In this way, the course of a scenario can be divided into several acts. The individual acts represent time periods of the scenario in which the actors involved perform specified activities. For the concept applied in this work, further conditions, for example on distance and relative speed of the actors, can optionally be included in the definition of an act. 

\paragraph{Activity}

An activity describes the change of state of an actor in a qualitative manner. E.\,g., for the identification of the scenario cut-in a purely lateral distinction of the activities is sufficient: lane change to the left, lane change to the right and lane following. This is the case as the longitudinal behaviour of the vehicles plays no role in the logical description of the scenario.

\paragraph{Event}

An event marks a point in time when one or more threshold values are reached, exceeded or undershot. Accordingly, a set of one or more conditions is specified for each event. As soon as such a set of conditions is fulfilled, the corresponding event is said to have occurred.

\subsection{Ontology for Driving Scenarios}\label{sec:ontology}

In this Section the developed ontology for the identification of driving scenarios in field data is presented. As in \parencite{Gelder.30.01.2020}, the ontology introduced in this work is formally represented by a domain model, which is briefly explained in the following. Afterwards, it is exemplarily shown how the driving scenario cut-in is defined using this domain model.

\begin{figure}[t!]
\centering\includegraphics[width=1.0\columnwidth]{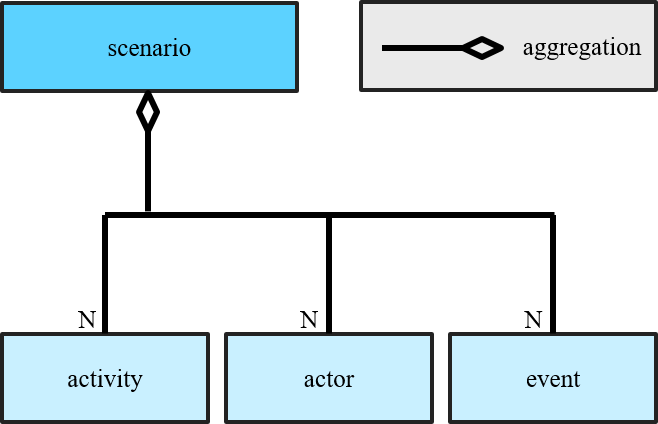}\caption{UML class diagram of the ontology for scenarios.}\label{fig:domain_model}
\end{figure}

\paragraph{Domain Model}
The classes of the domain model that are used to define a logical scenario are shown in \autoref{fig:domain_model}. The representation is based on the \textit{Unified Modeling Language}. Each of the blue blocks represents a class. The relationships between these different classes are described by the arrows. An arrow with a diamond symbol can be described as “is part of”. A “N” at the beginning of the arrow means that zero, one or more objects are part of an object of the class at the end of the arrow.

The ego vehicle and its dynamic environment are qualitatively described by objects of the classes activity, actor and event. One attribute of an object of the class scenario is the chronological list of acts. The acts describe which actor carries out which activity as well as which event conditions apply. It is possible that an actor carries out different activities and that an activity is carried out by different actors at the same time. 

\paragraph{Scenario Cut-in and its Attributes}
In the following, the benefits of the ontology are illustrated using the domain model for the definition of the cut-in scenario. To describe the scenario, it is necessary to define two events. The first one is the event following. The latter's conditions are met as soon as the ego vehicle follows a vehicle in the same lane within a specified maximum distance and within a specified maximum relative speed. The threshold value for the distance can be specified as a function of the driving speed of the ego vehicle. In general, a higher driving speed also requires interaction with vehicles that are further away. In addition, a threshold value that is independent of the driving speed can be defined, at which the distance condition of the subsequent driving event is met in every case. This ensures that the following event can be detected even when traffic is moving slowly. The second required event is called driving parallel. Its conditions are met as soon as a selected vehicle drives in one of the adjacent lanes in the detection area in front of or next to the ego vehicle.

\begin{figure}[t!]
\centering\includegraphics[width=1.0\columnwidth]{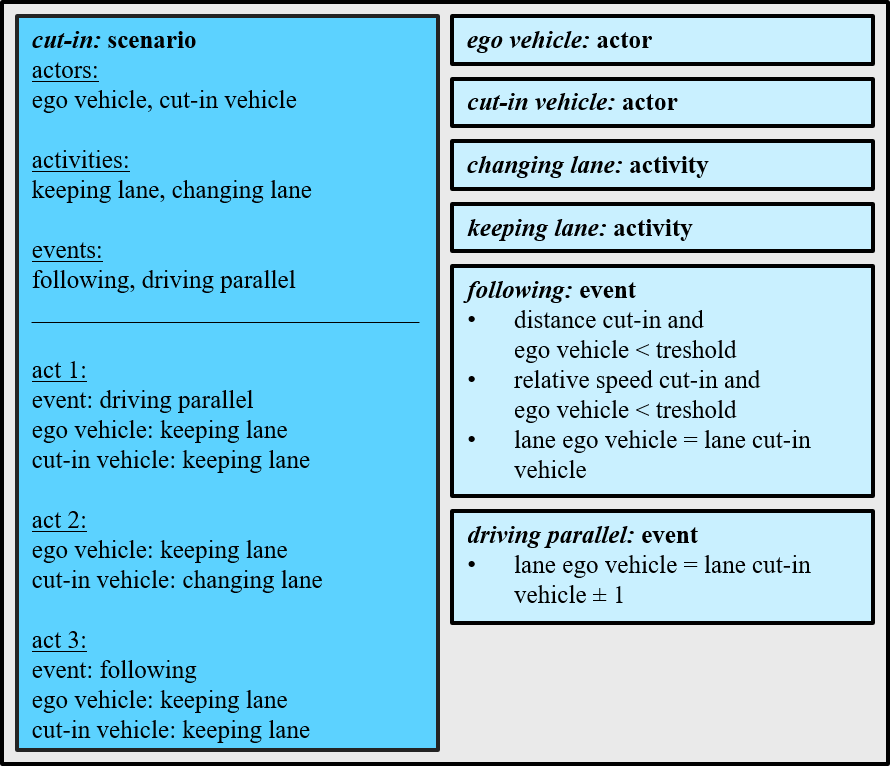}\caption{Objects for defining the scenario cut-in.}\label{fig:cut-in-model}
\end{figure}

To describe the cut-in scenario according to the presented domain model, objects are instantiated from the classes shown in \autoref{fig:domain_model}. \Autoref{fig:cut-in-model} shows the objects for the qualitative description of the scenario. The first line contains the object name in \textit{italics} and after the colon the name of the class from which the object was instantiated.

\subsection{Identification of Driving Scenarios}\label{sec:identification}

Objects in the measurement data are classified according to their relative position to the ego vehicle. A distinction is made between \textit{first-left}, \textit{first-right}, \textit{second-left}, \textit{second-right}, \textit{first-ego} and \textit{second-ego}. \textit{Left} in the designation means that the object is driving in the lane that is adjacent to the one of the ego vehicle on the left (equivalent for \textit{right}). \textit{Ego} means that the vehicle drives in the same lane as the ego vehicle. \textit{First} means that the vehicle is the first object detected in the respective lane in the direction of travel, \textit{Second} is the vehicle that drives in front of the \textit{First}.

For the identification of driving scenarios, an abstract description of field data is derived. The abstraction is based on the ontology explained in Section \ref{sec:ontology} with actors, their activities and events. By automatically comparing the measurement drives abstracted in this form with the scenarios previously described using the same elements, any concurrences can be recognised and the scenarios sought can thus be identified.

\paragraph{Abstraction of Measurement Data}

The identification of different events forms the basis for deriving the abstract description of a measurement drive from the field data. A distinction is made between two different types of events. On the one hand, there are the events according to the definition in Section \ref{sec:definitions}. On the other hand, there are the events that mark the points in time at which a status change takes place. The latter are only necessary for the practical derivation of the abstract description of a test drive and are not part of the definition of scenarios.

The events that mark a threshold value violation include the events following and driving parallel. The basis for the identification of the lane change activities of the ego vehicle and the other vehicles in its dynamic environment are formed by the lead change and ego lane change events. The latter mark times when a status change takes place. The lead change event marks the point in time at which the signal for the object ID of the first-ego vehicle changes. Accordingly, this also includes the times at which a vehicle driving directly ahead of the ego vehicle leaves the detection area or changes lanes to one of the adjacent lanes and there is no new driver ahead in the lane of the ego vehicle. The ego lane change event marks the point in time at which the ego lane assignment changes. Both events - lead change and ego lane change - occur in the case of a lane change when the corresponding vehicle is in the middle between two lanes.

Amongst others, possible activities to be identified for the derivation of an abstract description of the measurement drive are lane changes of the ego vehicle, lane keeping of the ego vehicle, cut-in lane change activities in front of the ego vehicle, lane keeping of a vehicle that later on cuts into the lane of the ego vehicle and lane keeping of a driver directly in front in the lane of the ego vehicle.

In principle, the following variants are conceivable for all lane change activities: the vehicle changes to one of the adjacent lanes, the vehicle changes over more than one lane or the vehicle only briefly switches to one of the adjacent lanes and then switches back again. In the given context, when identifying the time spans in which the relevant actors perform a lane change activity, a distinction is made between valid and invalid lane changes according to the conformity for the scenarios to be identified. If a vehicle changes from a common lateral position of one lane to a common lane position of another lane, a valid lane change activity is recognised for the time span in which the vehicle moves between these two positions. The change over more than one lane is also included, since a common lateral position on one of the originally adjacent lanes is also necessarily crossed. In the context of this work, a common lateral lane position is given when the center of a vehicle is more than 35 \% of the lane width away from the lane markings. If a vehicle only slightly overlaps one of the adjacent lane markings and consequently does not reach a common lateral lane position before it changes back to the original lane, an invalid lane change activity is recognised.

The activity lane keeping is recognised for all time spans in which the relevant actors do not carry out any valid or invalid lane change activity. In general, activities of the actors in the dynamic environment of the ego vehicle are only registered while the ego vehicle is performing the activity lane keeping. This is sufficient, since the continuous lane keeping activity of the ego vehicle is required for all scenarios to be identified.

\paragraph{Pattern Recognition}

\begin{figure}[t!]
\centering\includegraphics[width=1.0\columnwidth]{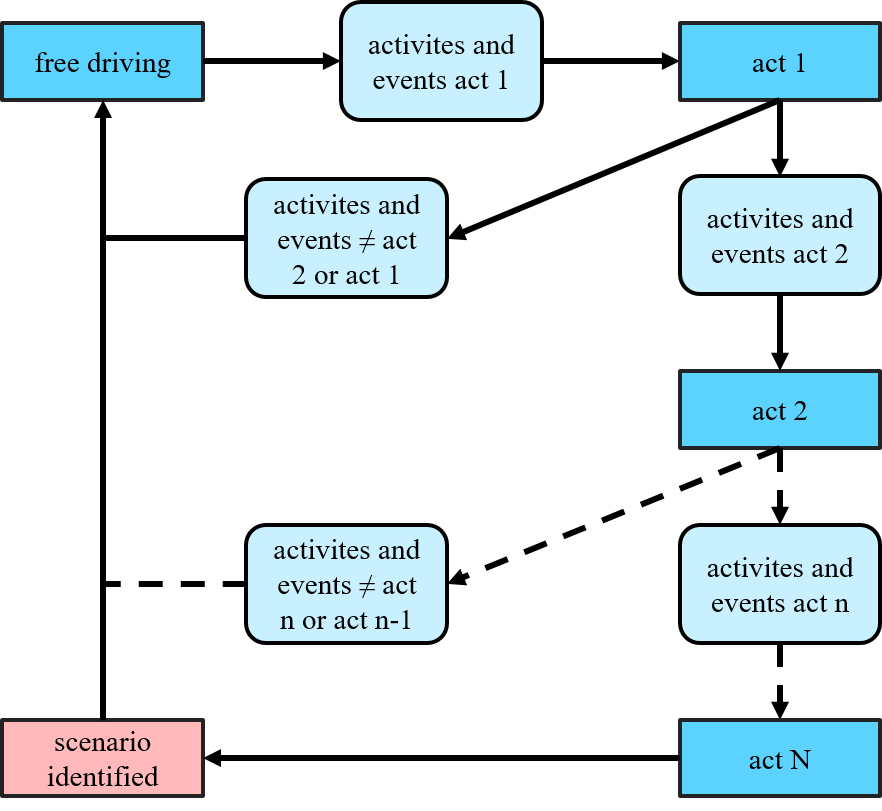}\caption{Scheme for pattern recognition.}\label{fig:pat_rec}
\end{figure}

For the identification of the scenarios sought, their descriptions according to the domain model are compared with the abstract descriptions derived from the field data. This comparison can always be carried out using the same procedure, regardless of the scenario (c.\,f. \autoref{fig:pat_rec}).

For identification, the derived activities and events are checked chronologically for each point in time of the measurement drive. The default status of the check is free driving - this means that no potential scenario has been identified. If the activities and events specified for the first act of a scenario are active at a point in time during this check, a potential scenario with the status of the first act is recognized. If these conditions for events and activities are violated in the further temporal iteration, the potentially recognized scenario is discarded if the changed activities and events do not correspond to those of the second act. In the latter case, the status is set to the second act. The same approach is used for the other acts of the scenario sought.

If a potentially recognized scenario is discarded during the iterations because the conditions for the current status and the subsequent act of a potentially recognized scenario are violated, the status free driving is set again. In the subsequent time steps, the conditions for events and activities of the first act are checked again.

If a potentially recognized scenario has the status of its last defined act (in \autoref{fig:pat_rec} act N), the scenario is noted as identified after a period of time specified for this act from the beginning of the first act to this point in time. If a maximum period of time has also been specified for the first act, the beginning of the scenario is set accordingly after the point in time at which the conditions of the first act were met for the first time.

\section{RESULTS AND DISCUSSION}\label{sec:results}

The exemplary implementation of the developed procedure for identifying driving scenarios was carried out in the object-oriented programming language Python. For its validation, a selection of measurement files, spanning one minute each, from test drives on highways was available together with video recordings. The aim of the validation was to check all relevant sub-functions and aspects for the identification of each potential scenario element for the scenario cut-in. For this purpose, a test catalogue was prepared, which specifies corresponding test cases for the identification of each event and activity. Suitable measurement files were assigned to these test cases by reviewing the video material. To carry out the validation, the measurement drive sections specified for the test cases were evaluated one after another using the implemented procedure. Afterwards, the evaluation was manually checked for false-positives and false-negatives using the video material according to the test specification.

All tests were performed for straight road sections. Since first trials have shown that vehicles overtaking on the outside of curves are sometimes mistakenly detected as first-ego and thus lead change events are detected by mistake. This is caused as the sensor system of the ego vehicle in our setting does not use the course of the road to classify the detected objects, but only their lateral offset to the ego vehicle. Similarly, it is not possible to determine the position of the detected objects in their lanes, since this type of information is not required for the function of the production vehicles.

Apart from such problems, which can be traced back to insufficient coverage of the vehicle environment, these inital investigations delivered a promising result. They show, a suitable approach for the identification of driving scenarios in field data could be elaborated, implemented and validated. The developed methodology with an universal ontology allows to reduce a large amount of measurement data to the relevant variables for scenario identification.

\subsection{Event Identification}

The identification of events worked without errors for the tests carried out on straight sections. The validation included both the events that mark points in time when one or more thresholds are violated, such as driving parallel and following, but also the events that mark times when a status change takes place, such as lead change and ego lane change.

The partially incorrect detection of lead change events in curves is not an error in the methodology. The requirement for the overall system is that the role assignment of the sensor system for the detected vehicles is correct. Thus, an object declared as \textit{first-ego} should actually drive ahead of the ego vehicle in the same lane. It is therefore obvious to directly address the cause in the sensor technology for the problem of false event identification. In addition to the relative position to the ego vehicle, the object detection layer should also take into account the position of the detected objects in relation to the lane markings. However, an optimisation to this effect has not been carried out at the time of publication of this work. If it should turn out that an adaptation is not possible, further considerations should be made how this problem can be circumvented and how the object classification can be corrected.

\subsection{Activity Identification}

During the identification tests, it has been shown that both valid and invalid lane changes of the ego vehicle are detected correctly.  However, a lane change that is not carried out into the lane centre area in front of the ego vehicle is incorrectly recognised as valid. The reason for this is that lane-related position information, for a procedure as described in Section \ref{sec:identification}, was only available for the ego vehicle in the used measurement data.

Vehicles driving ahead of the ego vehicle already have a high lateral offset to the ego vehicle when the road is slightly curved. This effect increases with increasing distance between the vehicles. Information on the course of the road and lane markings with which this effect could be calculated was not available for this work.

Instead, a simplified identification for lane change activities was implemented. Here, the time spans of all lane changes, independent of the actual duration, are determined by means of constant time intervals to the lead change event. Consequently, a lane change without reaching the lane centre area followed by a change back is falsely identified as valid. Similarly, the identified time span of the lane change does not always match the actual duration. Specifically, the described investigations were carried out with a duration of one second before and after the corresponding lane change event. The evaluation of video material showed that this duration mostly satisfactorily describes the real driving event for the application.

Despite this problem, the correct identification of ego vehicle lane changes shows that the developed method for identifying lane changes works. In principle, the implementation for the ego vehicle can be transferred to other vehicles equivalently, provided that required position signals are available in the future. In terms of the requirements of this work, this is a satisfactory result.

\subsection{Scenario Identification}

In the examinations carried out, the correct identification of driving scenarios was confirmed by the manual comparison with the identified activities and events. Specifically, the timing and overlap of these activities and events were checked against the pattern described for the scenario.

\subsection{Significance of the Results}

The achieved results provide a proof of concept for the functionality of the developed procedure for the identification of driving scenarios and the defined activities and events. However, the test cases only cover the driving situations that could be derived from the limited amount of available measurement data at the time of publication.

A further validation is recommended on the basis of a larger amount of measurement data for which the time periods of the scenarios contained are already marked correctly. Not least in order to be able to exclude possible errors in the identification due to special cases not taken into account with regard to the course of the road or the behaviour of other road users. In order to obtain a statistically reliable result for the application in system validation, the existing problems should be eliminated in advance by optimising the environment perception of the measuring vehicles.

\subsection{Discussion of the Methodology}

The presented method for metamodelling scenarios enables clear communication about their definition. The employed class structure also allows a simple and easy to understand transfer into an object-oriented software for the identification of scenarios in field data. Besides, it offers added value for the implementation of new scenarios to be identified. A new scenario instance can be created straight forward by referencing existing activities and events. The logic for identifying a scenario instantiated this way can be implemented universally and does not have to be adapted for new scenarios.
 
\begin{figure}[t!]
\centering\includegraphics[width=1.0\columnwidth]{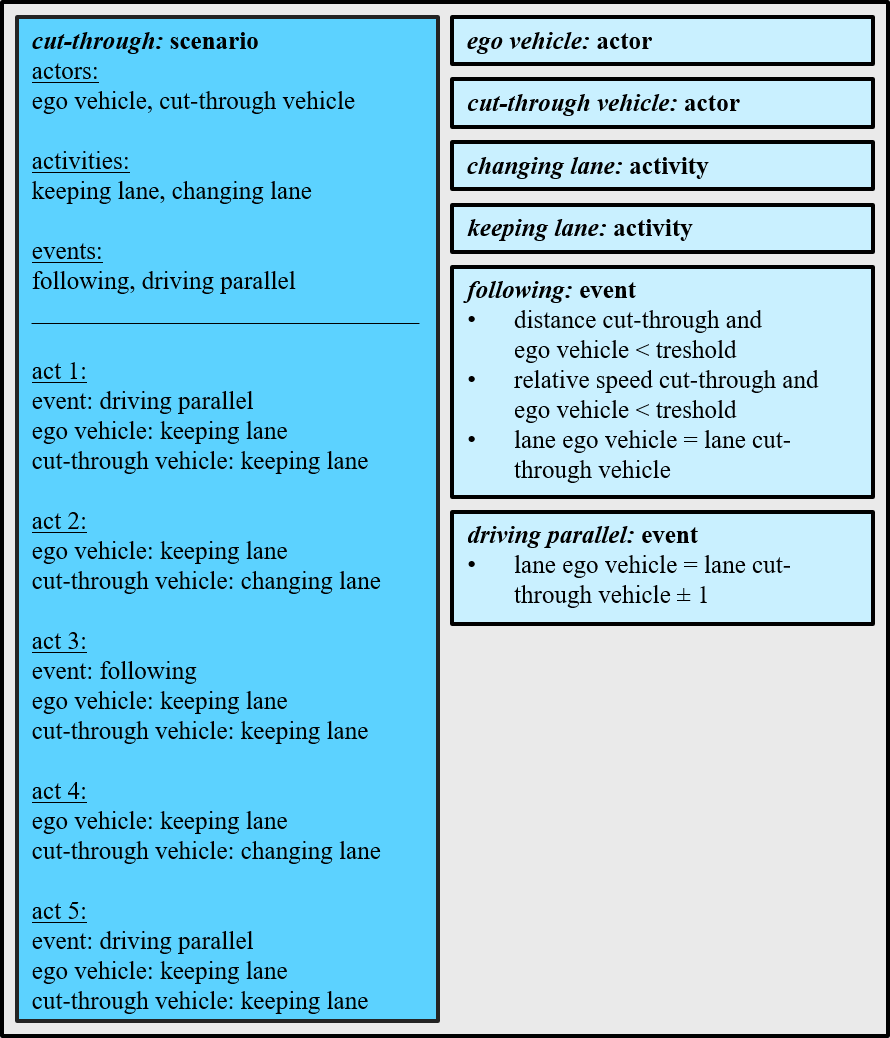}\caption{Objects for defining the scenario cut-through.}\label{fig:cut_through}
\end{figure}

Another relevant scenario for the highway domain is the so-called cut-through scenario. In the latter, a vehicle initially driving in a lane adjacent to the ego vehicle's, changes lanes ahead and into the lane of the ego vehicle and holds this lane for an undefined period of time. Afterwards, the vehicle changes lanes again in the same direction to the other adjacent lane and keeps it. The scenario can be instantiated as shown in \autoref{fig:cut_through} analogous to the scenario presented in Section \ref{sec:ontology}.

However, there are scenarios in the highway domain that require more detailed metamodelling, such as the lost-cargo or crossing-object scenarios. In these cases, the modelling must be extended to include objects lying on or moving across the street. Furthermore, corresponding signals must be used to derive them. Nevertheless, it should be possible to retain the presented concept for metamodelling with a domain model for such scenarios.	 
	 
\section{CONCLUSIONS}\label{sec:conclusion}

In this work, an approach enabling the identification of highway scenarios in field data was presented, exemplarily implemented and validated. The approach is based on qualitative scenario modelling. The modelled scenarios are detected in field data, which are abstracted beforehand. To the best of our knowledge, a detailed description of the actual process of pattern recognition and its application to real-world data has not yet been published. The results obtained in first experiments are promising and provide a proof of concept for the approach's functionality. Nevertheless, further investigations based on a larger amount of measurement data will be carried out after solving shortcomings in the measurement technology.

In addition to an even more detailed modelling of the scenarios, the identification algorithm could be extended with machine learning methods in the future. Such methods would require a large amount of learning data in which the scenarios sought are clearly identified.

\printbibliography

\end{document}